\documentclass[conference]{IEEEtran}
\IEEEoverridecommandlockouts
\usepackage{cite}
\usepackage{amsmath,amssymb,amsfonts}
\usepackage[colorlinks,linkcolor=magenta]{hyperref}
\usepackage{algorithmic}
\usepackage{graphicx}
\usepackage{textcomp}
\usepackage{makecell}
\usepackage{xcolor}
\usepackage{bbding}
\usepackage{pifont}

\def\BibTeX{{\rm B\kern-.05em{\sc i\kern-.025em b}\kern-.08em
    T\kern-.1667em\lower.7ex\hbox{E}\kern-.125emX}}
\begin{document}

\title{AutoKary2022: A Large-Scale Densely Annotated Dataset for Chromosome Instance Segmentation}

\author{\IEEEauthorblockN{1\textsuperscript{st} Dan You}
\IEEEauthorblockA{\textit{Hangzhou City University} \\
Hangzhou, China \\
961031769@qq.com}
\and
\IEEEauthorblockN{2\textsuperscript{nd} Pengcheng Xia}
\IEEEauthorblockA{\textit{Hangzhou City University} \\
Hangzhou, China \\
 2835162281@qq.com}
\and
\IEEEauthorblockN{2\textsuperscript{nd} Qiuzhu Chen}
\IEEEauthorblockA{\textit{Hangzhou City University} \\
Hangzhou, China \\
995429302@qq.com}
\and
\IEEEauthorblockN{3\textsuperscript{rd} Minghui Wu}
\IEEEauthorblockA{\textit{Hangzhou City University} \\
Hangzhou, China \\
mhwu@zucc.edu.cn}
\and
\IEEEauthorblockN{ Suncheng Xiang$^*$}
\IEEEauthorblockA{\textit{Shanghai Jiao Tong University} \\
Shanghai, China \\
xiangsuncheng17@sjtu.edu.cn}
\and
\IEEEauthorblockN{Jun Wang$^*$}
\IEEEauthorblockA{\textit{Hangzhou City University} \\
Hangzhou, China \\
wjcy19870122@163.com}
}

\maketitle

\begin{abstract}
Automated chromosome instance segmentation from metaphase cell microscopic images is critical for the diagnosis of chromosomal disorders (\textit{i.e.}, karyotype analysis). However, it is still a challenging task due to lacking of densely annotated datasets and the complicated morphologies of chromosomes, \textit{e.g.}, dense distribution, arbitrary orientations, and wide range of lengths. To facilitate the development of this area, we take a big step forward and manually construct a large-scale densely annotated dataset named \textbf{\textit{AutoKary2022}}, which contains over 27,000 chromosome instances in 612 microscopic images from 50 patients. Specifically, each instance is annotated with a polygonal mask and a class label to assist in precise chromosome detection and segmentation. On top of it,
we systematically investigate representative methods on this dataset and obtain a number of interesting findings, which helps us have a deeper understanding of the fundamental problems in chromosome instance segmentation. We hope this dataset could advance research towards medical understanding. The dataset can be available at:\href{https://github.com/wangjuncongyu/chromosome-instance-segmentation-dataset}{https://github.com/wangjuncongyu/chromosome-instance-segmentation-dataset}.
\end{abstract}


\section{Introduction}

A healthy human cell has 23 pairs of chromosomes, including the initial 22 pairs of autosomes and the 23rd pair of sex chromosomes (\textit{i.e.}, the XY and XX stand for male and female, respectively). Besides, every normal chromosome has certain structures, \textit{e.g.}, the centromere positions and the ratios between the long and short arms~\cite{McGowan-Jordan_Simons_Schmid_2016}. Any numerical or structural variations of chromosomes may cause disorders such as intellectual disability, congenital malformations, sterility, sexual variations, and even cancers~\cite{2010Disorders}. In clinical practice, karyotype analysis~\cite{https://doi.org/10.1111/exsy.12799} is a routine procedure for the detection and diagnosis of the chromosomal disorders.

\begin{table*}[!t]
\begin{center}
\caption{ The results of other classic methods.}
\label{tab1}
\setlength{\tabcolsep}{3.5mm}{
\begin{tabular}{l|c|c|c}
  \Xhline{0.8pt}
  Study & Method & Task & Performance
  \\
  \hline
  Minaee \textit{et al.}~\cite{7163174} & Traditional method & Segmentation & Accuracy=91.9\% (binary class) \\
  Yilmaz \textit{et al.}~\cite{8599328} & YOLOv2 & Detection & AP = 99.2\% (binary class) \\
  P. Wang \textit{et al.}~\cite{9630695} & Rotated Mask R-CNN & Segmentation & AP@50=95. 8\% (binary class);
mAP@50=65. 9\% (multi class) \\
  \Xhline{0.8pt}
\end{tabular}}
\end{center}
\end{table*}

\begin{table*}[!t]
\begin{center}
\caption{Comparisons With existing datasets. ``Raw data" represents MC image without any processing, and ``Public" indicates the dataset is open source, respectively.} \label{tab2}
\setlength{\tabcolsep}{3.5mm}{
\begin{tabular}{l|c|c|c|c|c|c}
  \Xhline{0.8pt}
  Methods & train images & test images & annotations & resolution & Raw data & Public
  \\
  \hline
  Minaee \textit{et al.}~\cite{7163174} & - & - & 62 & - & \textcolor{black}{\ding{56}}  & \textcolor{black}{\ding{56}}  \\
  Yilmaz \textit{et al.}~\cite{8599328} & 145 & - & 6678 & - & \textcolor{black}{\ding{56}} & \textcolor{black}{\ding{56}} \\
  Al-Kharraz \textit{et al.}~\cite{9178721} &  118 & 29 & - & $910\times910 $ & \textcolor{black}{\ding{56}} & \textcolor{black}{\ding{56}} \\
  P. Wang \textit{et al.}~\cite{9630695} &  1103 & 137 & - & $640\times512$ & \textcolor{black}{\ding{56}} & \textcolor{black}{\ding{56}} \\
  \hline
  \textbf{AutoKary 2022 (Ours)} & 547 & 67 & 27109 & $3200\times2200$ & \textcolor[rgb]{0.00,0.00,0.00}{\ding{52}} & \textcolor[rgb]{0.00,0.00,0.00}{\ding{52}}\\
  \Xhline{0.8pt}
\end{tabular}}
\end{center}
\end{table*}

\begin{figure}[!t]
\centering
\centerline{\includegraphics[width=0.9\linewidth]{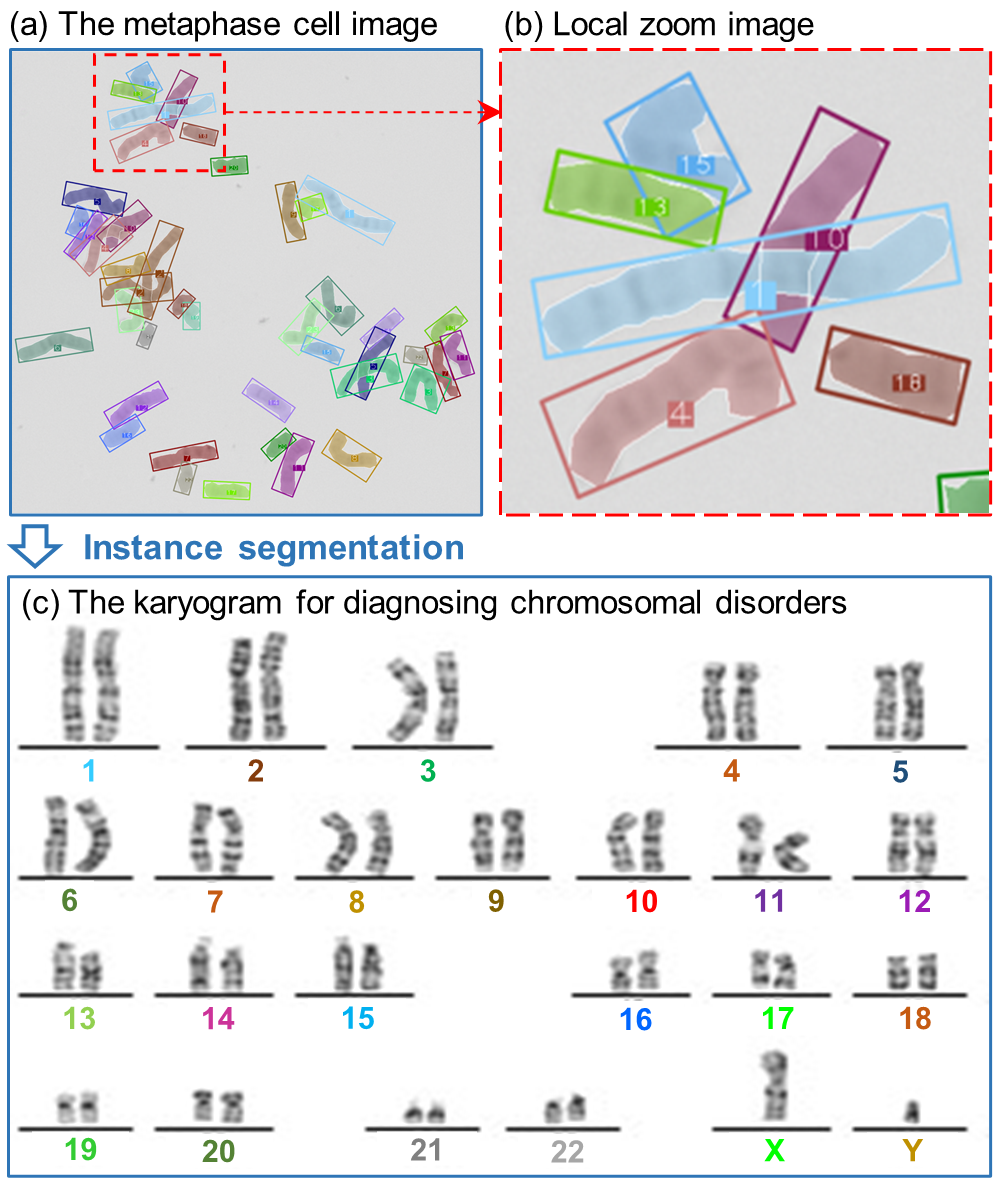}}
\caption{The layout of karyogram from metaphase cell image, which can be treated as an instance segmentation problem. However, it is a challenging task due to the complicated features of chromosomes such as dense distribution, arbitrary orientations, wide range of lengths, bend and slender shapes.}
\label{fig1}
\end{figure}

Although karyotype analysis has a long history of more than 50 years, it still suffers from some pain points. For example, as illustrated in Fig.~\ref{fig1}, a key step of karyotype analysis is the layout of a so-called karyogram from metaphase cell (MC) images that are acquired via micro-scanning Giemsa-stained cells of patients. Cytogeneticists observe the structures and features of chromosomes presented on the karyogram and finally report results. However, the layout of karyogram requires experienced technicians to manually count, extract and classify chromosome instances from the MC images. This process is extremely labor-intensive and time-consuming. It is reported that the processing time for each case is about 30-50 minutes~\cite{9134813}. Consequently, automated layout of the karyogram is an urgent need to improve the efficiency and reduce the costs of the examination.

In practice, the layout of karyogram can be treated as an instance segmentation problem ~\cite{9630695}, which simultaneously detects, segments and classifies objects from an image using a single model. Although many deep-learning-based methods such as the Mask R-CNN~\cite{He_2017_ICCV} and its variants~\cite{8917599,8953609,Chen_2019_CVPR} have been developed for instance segmentation in natural images, reliable chromosome instance segmentation still faces some challenges. For example, as demonstrated in Fig.~\ref{fig1}, the chromosomes have distinctive features that are quite different from objects in natural or other medical images, including dense distribution (even overlapped), arbitrary orientations, wide range of lengths, bend and slender shapes, and other complicated situations as well. Existing methods may fail to cope with these situations due to their limited discriminability. Importantly, training deep learning models requires large-scale labeled segmentation datasets. However, as far as we know, this research community still lacks such datasets for chromosome instance segmentation. Specially, previous chromosome datasets either contain overlap situation of chromosome contacts in many synthetic samples, or not released to the public, which makes it infeasible to quantitatively analyze the performance of different segmentation methods based on a unified standard.

\begin{figure}[!t]
\centerline{\includegraphics[width=0.9\linewidth]{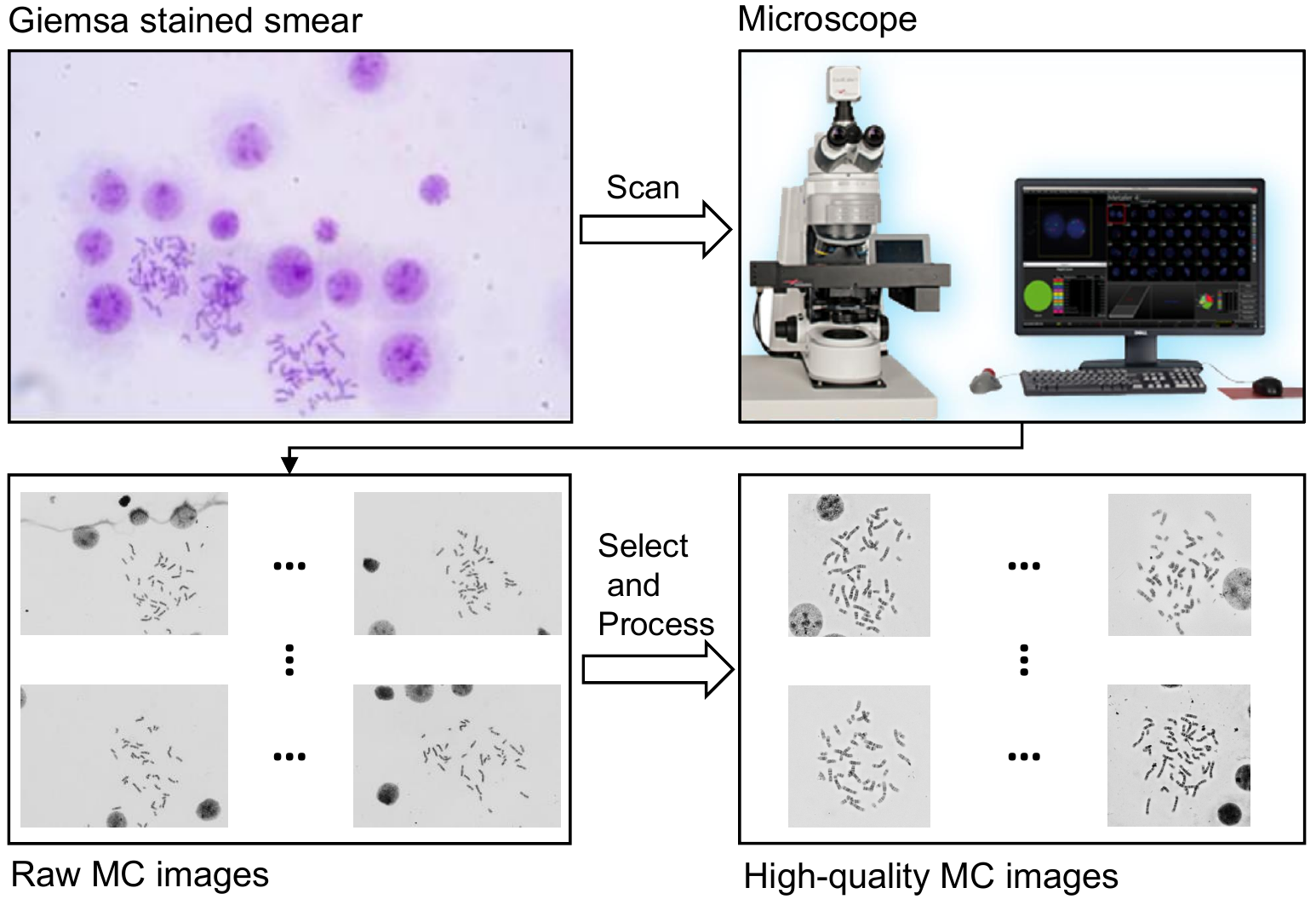}}
\caption{The process of data collection.}
\label{fig2}
\end{figure}

\begin{figure*}[!t]
\centerline{\includegraphics[width=0.97\linewidth]{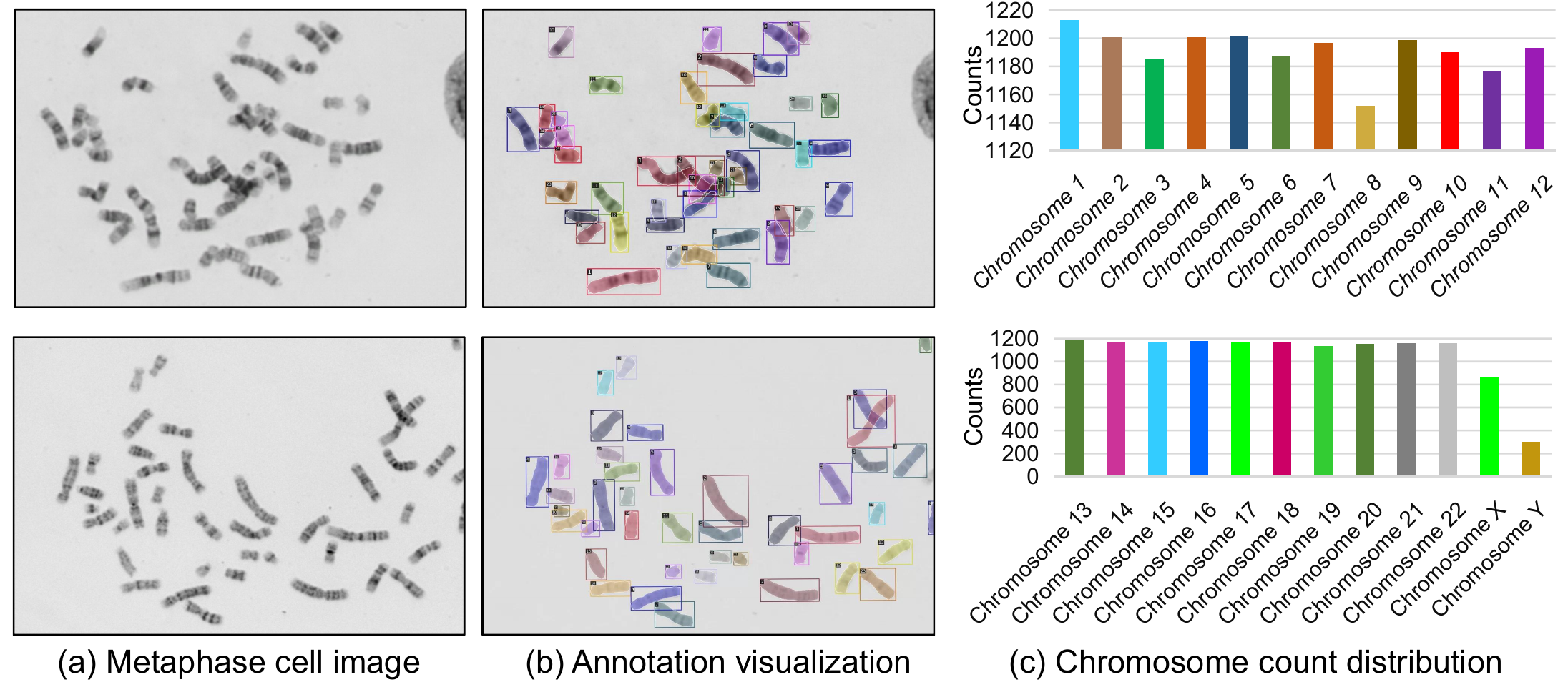}}
\caption{(a) Two representative MC images. (b) Ground truth visualization image. (c) The number of each chromosome category in the dataset.}
\label{fig3}
\end{figure*}

To address these challenges mentioned above, in this work, we manually construct a large-scale densely annotated dataset named \textit{\textbf{AutoKary2022}} for chromosome instance segmentation, which contains over 27,000 chromosome instances in 612 microscopic images from 50 patients. Based on it, we perform quantitative study and evaluate the performances of recently proposed instance segmentation models, \textit{e.g.}, SOLOv2~\cite{DBLP:journals/corr/abs-2003-10152}, PolarMask++~\cite{DBLP:journals/corr/abs-2105-02184}, Mask R-CNN~\cite{He_2017_ICCV}, Cascade R-CNN~\cite{8917599}, SCNet~\cite{9156634} , HTC~\cite{Chen_2019_CVPR}, and MS R-CNN~\cite{8953609}. Comprehensive Experiments conducted on our dataset not only lead to a number of attractive findings that are beneficial for future research, but also clearly show new challenges posed by \textit{AutoKary2022}.


Overall, our work contributes to the research of chromosome instance segmentation in two different ways: 1) We manually construct a large-scale densely annotated dataset named \textit{AutoKary2022}, which provides high-quality and densely annotated annotations. In particular, the annotations are applicable for multi-task learning. 2) We conduct in-depth study on top of \textit{AutoKary2022}, which reveals the key challenges that arise in the densely annotated setting, which may point to new directions of future research.

\section{Related Works}

\subsection{Methods of Chromosome Segmentation}
Early chromosome instance segmentation was mainly based on statistics and geometry, which adopts threshold, edge, region and other related technologies to achieve better performance for image semantic segmentation~\cite{https://doi.org/10.1111/exsy.12799}. For instance, N.Madian \textit{et al.}~\cite{6347213} propose an algorithm to separate the touching chromosomes and the segmentation of overlapping chromosomes from G-Band metaspread images. E.Grisan \textit{et al.}~\cite{4773198} show a local adaptive threshold method. They divide the image in a tessellation of squares of fixed dimension and  evaluated the Otsu~\cite{otsu1979threshold} threshold of each square separately. S. Minaee \textit{et al.}~\cite{7163174} introduce a geometric-based method which is used for automatic detection of touching and overlapping chromosomes and separating them. Unfortunately, traditional methods always fail to produce competitive results due to its limited discriminability when dealing with the complex situation of chromosomes.

Recently, deep learning based methods have tremendously pushed forward the boundary of chromosome instance segmentation performance. These methods focus on designing various deep CNN structures to learn discriminative feature embeddings and/or strive to devise better loss functions for training the network.
For example, Yilmaz \textit{et al.}~\cite{8599328} introduce an end-to-end framework which can process chromosome images and identify individual chromosomes and clusters of chromosomes. 
Al-Kharraz \textit{et al.}~\cite{9178721} propose a deep learning-based chromosome detection and classification system, where a single chromosome is detected and processed using a YOLOv2 convolutional neural network.
P. Wang \textit{et al.}~\cite{9630695} proposed an enhanced rotated Mask R-CNN\cite{He_2017_ICCV} method to detect axis-aligned and rotated bounding boxes, enabling the network to combine the advantages of axis-aligned and rotated bounding boxes for chromosome detection in chromosome segmentation. Table~\ref{tab1}  shows the results of these studies. Because different studies focused on different tasks and utilized diverse datasets and metrics for evaluation, it is hard to compare these methods fairly. However, from the above table, it can be concluded that multi-class segmentation or detection of chromosomes (\textit{i.e.}, identifying 24 categories) is much challenging than binary segmentation or detection (regardless the chromosome’s category). For example, P. Wang \textit{et al.} achieved an AP score of 95.8\% for binary-class segmentation, but the counterpart is only 65.9\% for multi-class segmentation. Even though these studies can obtain pleasing results, the performance might be still limited because their approches were mainly developed for natrual images.  There is a large gap between natrual images and chromosme dataset. Importantly, some of those approaches requires a very large dataset to train the model properly so they may not be applicable to small datasets, because they could be very prone to over-fitting~\cite{xiang2021less}.




\subsection{Chromosome Instance Dataset}
As for the dataset of chromosome instance segmentation, previous research works~\cite{6347213,7163174} always consist of a small number of individually touching or overlapping chromosome instances. For example,
Yilmaz \textit{et al.}~\cite{8599328} present proposed an automated method for segmenting and separating G segments of human chromosomes. Al-Kharraz \textit{et al.}~\cite{9178721} propose a system to detect individual chromosomes and classify them using deep learning techniques. Unfortunately,
the above chromosome datasets are either unextensible or not released to the public, it is impossible to reproduce the experimental results or confirm whether it has actual clinical application value.
In contrast, our \textit{AutoKary2022} dataset contains complete 23 pairs of chromosomes, which is the original MC images without any clinical manual processing and preserves the original medical knowledge. Table~~\ref{tab2} summarizes the main differences between our dataset and existing datasets. To the best of our knowledge, this is the first chromosome instance dataset with densely annotation on segmentation task. which can advance the research for medical understanding significantly.

\section{The Proposed AUTOKARY2022 dataset}

In this section, we describe \textit{AutoKary2022}, a large-scale dataset with high-quality annotations to medical community. Below we at first review the process of constructing and
annotation collection, then present an analysis over dataset statistics.

\begin{figure}[!t]
\centerline{\includegraphics[width=0.75\linewidth]{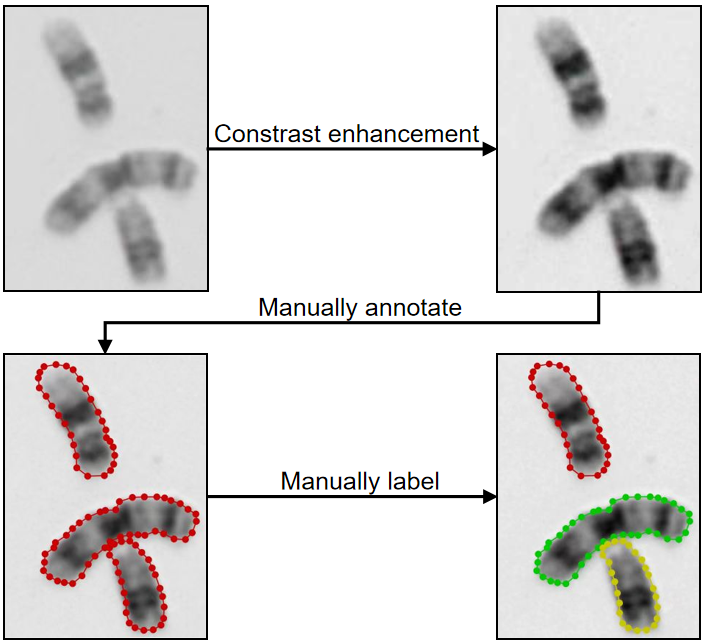}}
\caption{Chromosome labeling process.}
\label{fig4}
\end{figure}

\begin{figure*}[!t]
\centerline{\includegraphics[width=0.96\linewidth]{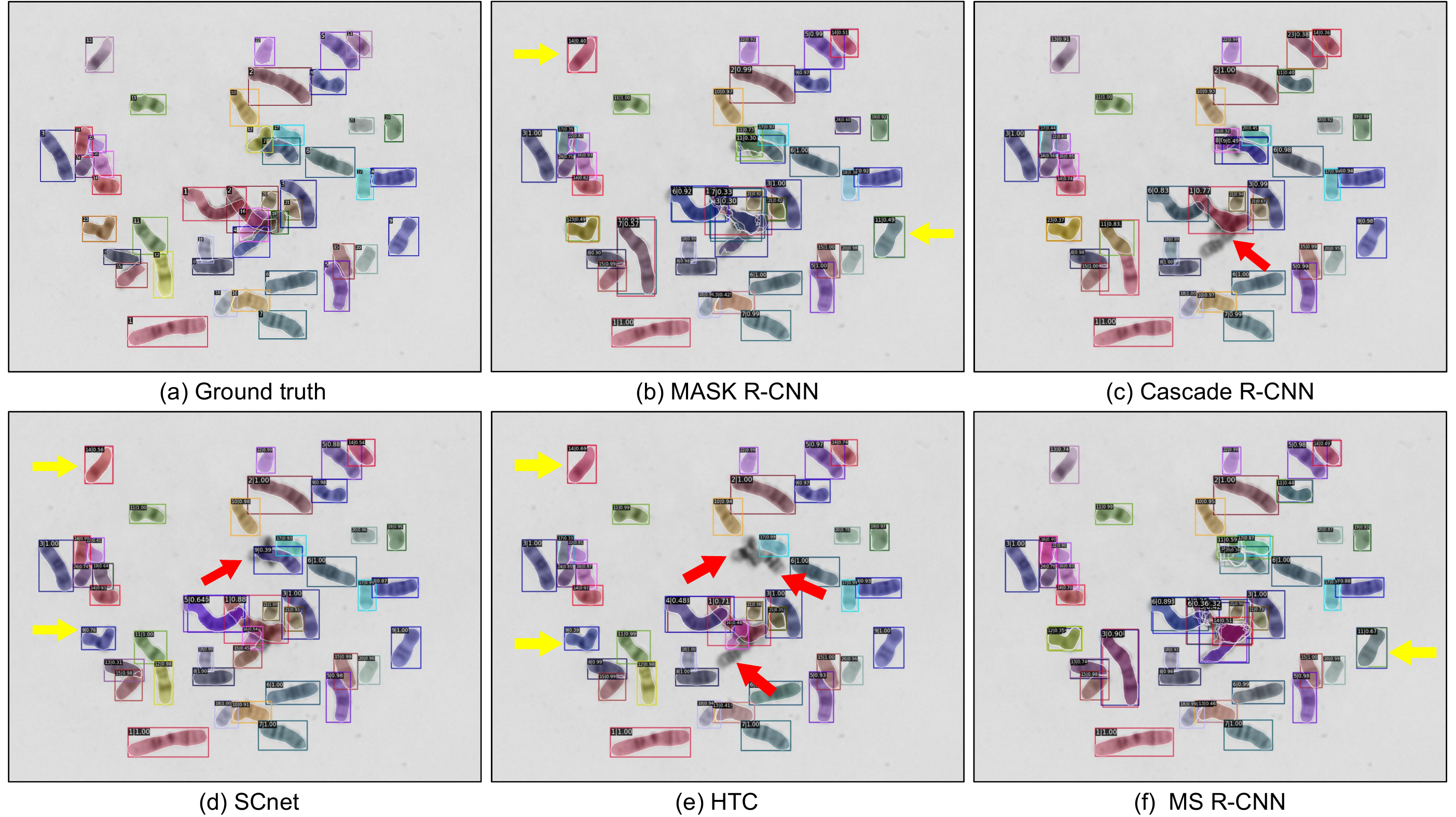}}
\caption{The visualisation of ground truth and predict box via various baselines. \textcolor[rgb]{0.80,0.80,0.00}{Yellow} arrows indicate misclassified chromosomes, \textcolor{red}{red} arrows indicate missed chromosomes.}
\label{fig5}
\end{figure*}

\subsection{Data Construction}
To train the chromosome instance segmentation model, we co-operate with a Hospital and collect microscopic images of metaphase cells from 50 patients. Using G-banding technology, each chromosome has a relatively constant banding feature so that we can accurate identification of chromosomes and the discovery of subtle structural aberrations on chromosomes. The data collection process is illustrated in Fig.~\ref{fig2}. We use the labelme~\cite{russell2008labelme} tool to label the dataset, which contains a total of 612 chromosome image samples, where each image has a resolution of $3,200\times2,200$ to facilitate training the model. The chromosome instance in each image has its corresponding annotation file, which includes bounding boxes and chromosome categories. Fig.~\ref{fig3} (a-b) depicts an MC image of chromosome instance annotation and its visualization of ground truth. From it we can observe that an MC image contains bent, interconnected, and overlapping chromosomes, which are in line with the characteristics of chromosomes in reality.

\subsection{Dataset Statistics}
\textit{AutoKary2022} has 50 folders (each fold corresponds to a patient), containing 612 MC images and a total of 27,109 chromosome instances. The count distribution of all chromosomes is plotted in Fig.~\ref{fig3} (c).The ratio of male to female patients was 1.5:1. The dataset is divided into train dataset and test dataset.
\textit{AutoKary2022}, on the other hand, is constructed through a large amount of annotation work. In contrast to existing chromosome datasets, it is built using clinical chromosome data without any artificial simulations. Our dataset instance is more complete,
 \textit{e.g.}, a MC image contains multiple pairs of chromosome instances, each of which has a corresponding classification label. This can be used for the training of chromosome instance segmentation models and chromosome classification models.

\subsection{Dataset Properties}
\textit{AutoKary2022} has several attracting properties that can bring great benefits to the community.

\textbf{Authenticity.} Unlike other datasets, \textit{AutoKary2022} retains the background and cellular residues and its exemplary distribution of chromosomes is not by later typesetting. This means that \textit{AutoKary2022} fits perfectly into the original morphology of the chromosome.

\textbf{High quality.} The \textit{AutoKary2022} labelling process is shown in Fig.~\ref{fig4}. Due to the non-rigid structure of the chromosomes, the length, width and specific shape of the chromosomes are deformed. They are hard to identify solely from their G-band patterns. In particular, some specific chromosome instances must be classified by exclusionary and comparative methods. In addition, the initial manual delineation is subject to problems such as misclassification of overlapping regions and misidentification of cellular residues as chromosomes. The labelling process is therefore a continuous cycle of validation of the above steps, which ensures that our dataset is highly accurate.

\textbf{Scalability.} Our \textit{AutoKary2022} dataset is extendable and can bring more benefits to the community. For example, it can be edited/extended not only for segmentation, but also for hromosomal object detection and instance
classification.

\section{Experiments}

\subsection{Experiment setup}
Using the \textit{AutoKary2022} dataset, we trained the Mask R-CNN~\cite{He_2017_ICCV} and its four state-of-the-art variants, including the Cascade R-CNN~\cite{8917599}, SCNet~\cite{9156634}, HTC~\cite{Chen_2019_CVPR}, and MS R-CNN~\cite{8953609} that were implemented in the MMDetection framework~\cite{chen2019mmdetection}. All models (with the same input size of
 $1,600\times800$ and the backbone network of ResNet-50) were trained on NVIDIA RTX3090 GPUs using the Adam optimizer with a learning rate of 0.0001, decaying every epoch with an exponential rate of 0.96. The total number of epochs was 50, and the batch size was 1. The mean Average Precision (mAP), mean Panoptic Quality (mPQ) and Aggregated Jaccard Index (AJI) were adopted to evaluate the chromosome instance segmentation performance of the models.

\begin{figure}[!t]
\centerline{\includegraphics[width=0.95\linewidth]{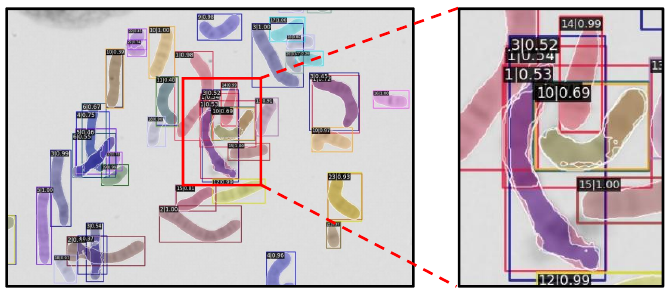}}
\caption{Mask R-CNN has limited performance at densely distributed chromosomes.}
\label{fig6}
\end{figure}

\begin{table}[!t]
\begin{center}
\caption{Performance comparison between different deep learning models on \textit{AutoKary2022} dataset. Measured by \%.}
\label{tab3}
\setlength{\tabcolsep}{3.5mm}{
\begin{tabular}{l|c|c|c}
\Xhline{0.8pt}
  Baseline & mAP@50 & mPQ & AJI
  \\
  \hline
  SOLOv2~\cite{DBLP:journals/corr/abs-2003-10152} & 83.3 & 71.5 & 50.7 \\
  PolarMask++~\cite{DBLP:journals/corr/abs-2105-02184} & 82.7 & 69.9 & 54.3 \\
  Mask R-CNN~\cite{He_2017_ICCV} & 90.3 & 83.8 & 70.5 \\
  Cascade R-CNN~\cite{8917599} & 93.0 & 84.7 & 73.1 \\
  SCNet~\cite{9156634} & 92.4 & 83.4 & 73.7 \\
  HTC~\cite{Chen_2019_CVPR} & 92.5 & 85.0 & 73.5 \\
  MS R-CNN~\cite{8953609} & 89.7 & 84.2 & 69.5 \\
\Xhline{0.8pt}
\end{tabular}}
\end{center}
\end{table}

\subsection{Evaluations}
The results are summarized in Table~\ref{tab3}, which demonstrates that the two-stage methods (with a subsequent feature refinement stage) can achieve much superior performance with the mAP@50 larger than 89\% to the single-stage methods (i.e., the SOLOv2~\cite{DBLP:journals/corr/abs-2003-10152} and the PolarMask++~\cite{DBLP:journals/corr/abs-2105-02184}). Promisingly, the variants, except the MS R-CNN~\cite{8953609}, are superior than the Mask R-CNN~\cite{He_2017_ICCV}. For example, the mAP@50 of Cascade R-CNN~\cite{8917599} is up to 93.0\%, with an improvement of 3.3\% compared to 89.7\% that achieved by the Mask R-CNN~\cite{He_2017_ICCV}.

\begin{figure}[!t]
\centerline{\includegraphics[width=1\linewidth]{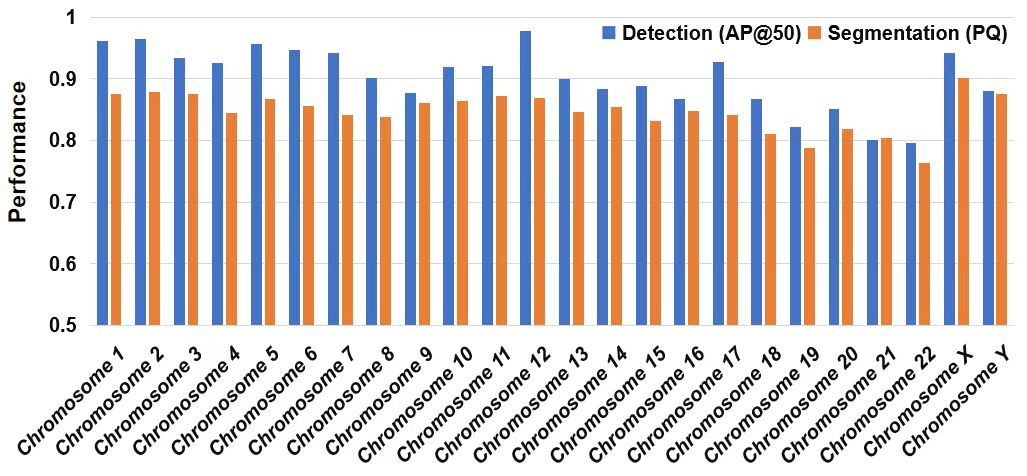}}
\caption{The AP of chromosomal object detection and the PQ of instance segmentation by Cascade R-CNN.}
\label{fig7}
\end{figure}

To analysis the results more intuitively, we visualize the predictions of all testing images. Fig.~\ref{fig5} shows the results of a representative case, and two main findings can be drawn. First, all methods can identify the independent chromosomes (\textit{i.e.}, the instances that are not connected or overlapped with others) well. However, there are still some misclassifications, especially for the short chromosomes such as the type 9, 13, and 23 as indicated by the yellow arrows in Fig.~\ref{fig5}. To further analysis this phenomenon, we plot the average precision (AP) and Panoptic Quality (PQ) of all categories that achieved by the Cascade R-CNN~\cite{8917599} in Fig.~\ref{fig7}. It can be observed that the APs and PQs of most short chromosomes, \textit{e.g.}, the types 9 to 22, are indeed inferior to that of other instances, We suspect that this is due to there is inter-class similarity between these chromosomes, \textit{e.g.}, the chromosomes of type 19 and 21 have very short length (small size) and similar G-band patterns (see Fig.~\ref{fig1}). Consequently, it is hard to learn discriminative features for distinguishing these chromosomes.; Second, Fig.~\ref{fig6} also reveals that existing methods have extremely poor performance in cluster regions where chromosomes are densely distributed. For instance, some chromosomes are mis-recognized as wrong types or even not identified as indicated by the red allows in Fig.~\ref{fig5}. Besides, we also provide an example about the prediction of cluster regions in Fig.~\ref{fig6}, from which we can observe that the predicted scores are relatively small (\textit{e.g.}, only with an accuracy of 0.52), and the predicted mask is far from satisfactory.

\subsection{Discussion}
According to the experimental results, our \textit{AutoKary2022} dataset can be successfully utilized to train models for chromosome instance segmentation in metaphase cell images. To go even further, we gave an explanation about two interesting phenomena observed during the experiment.

Firstly, as depicted in Fig.~\ref{fig5} and ~\ref{fig6}, we find that existing methods have poor performance in cluster regions. The root cause of this phenomenon might be that these methods predict bounding boxes from a series of horizontal anchors (\textit{i.e.}, predefined boxes) to identify each instance. Since the chromosomes have slender shapes and arbitrary orientations, a single anchor usually contains several intact or fragmentary instances in the cluster regions (see Fig.~\ref{fig6}), which may confuse the model from learning discriminative features and impede the subsequent procedures, \textit{i.e.}, segmentation and classification.

Secondly, as illustrated in Fig.~\ref{fig7}, there exists an obvious difference in terms of the segmentation performance among different chromosome categories. In particular, the performance of short chromosomes (\textit{i.e.}, from type 9 to 22) is much inferior to the long chromosomes. We attribute this phenomenon to the inter-class similarity between the short chromosomes with similar shapes and patterns (see Fig.~\ref{fig1}). It is difficult to learn discriminative features for distinguishing these chromosomes. On the other hand, existing methods follow the high-to-low (encoder) and low-to-high (decoder) feature extraction paradigm. This classic paradigm may lose detailed spatial information due
to the sampling operations~\cite{DBLP:journals/corr/abs-1902-09212}, which might further aggravate the situation. The above
challenges warrant further research and consideration when deploying our
\textit{AutoKary2022} dataset in real scenarios.

\section{Conclusion}
In this paper, we introduce a large-scale densely annotated chromosome dataset named \textit{AutoKary2022}, which contains several properties in multiple aspects, \textit{e.g.} Authenticity, High accuracy. On top of \textit{AutoKary2022}, we have empirically investigated representative methods on various task, such as chromosomal object detection and instance segmentation. These studies not only lead to a number of attractive findings that are beneficial for future research, but also clearly show new challenges posed by \textit{AutoKary2022}. We hope these efforts could facilitate new advances in the field of medical understanding.

\section{acknowledgment}
This research was supported in part by National Natural Science Foundation of China under Grant 62101318, National Student Innovation and Entrepreneurship Project under Grant 201000582673, and Key Research and Development Program of Jiangsu Province under Grant BE2020762. It was also supported by the advanced computing resources provided by the Supercomputing Center of Hangzhou City University.

\bibliographystyle{IEEEbib}
\bibliography{icme2023template}

\vspace{12pt}

\end{document}